%% file: main.tex
\documentclass
[
    conference,
    letterpaper
]
{IEEEtran}
\IEEEoverridecommandlockouts
\IEEEpubid{\makebox[\columnwidth]{\copyright2020 IEEE \hfill} \hspace{\columnsep}\makebox[\columnwidth]{ }}

\usepackage{graphicx}
\usepackage{subcaption}
\usepackage{booktabs}
\usepackage{multirow}
\graphicspath{{Images/}}
\usepackage[acronym]{glossaries}

\makeglossaries
\input{acronyms.tex}
 
\usepackage{xcolor}

\begin{document}

\title
{
    % Revisions
    % 1. Single-Neuron Temporal Feature Detection\\ Using Synaptic Delays\\in a Dynamic Neuromorphic Processor
    % 2. Spatiotemporal Feature Detection by Single Neurons in a Dynamic Neuromorphic Processor
    % 3. Spatiotemporal Feature Detection\\ with Single Neurons\\ in a Dynamic Neuromorphic Processor
    Synaptic Integration of Spatiotemporal Features\\
    with a Dynamic Neuromorphic Processor
    \thanks
    {
        This work is funded by The \mbox{Kempe} Foundations,
        under contracts JCK-1809 and SMK-1429,
        and by ECSEL JU,
        under grant agreement no.\ 737459.
        %and is based on a collaboration with the Institute of \mbox{Neuroinformatics} supported by STINT under contract IG2011-2025.
    }
}

\author
{
    \IEEEauthorblockN
    {
        Mattias~Nilsson,
        Foteini~Liwicki,
        and Fredrik~Sandin
    }
    \IEEEauthorblockA
    {
        \textit{Embedded Intelligent Systems Lab (EISLAB)}\\
        \textit{Lule{\aa} University of Technology, 971 87 Lule{\aa}, Sweden}\\
        mattias.1.nilsson@ltu.se
        %, \{foteini.liwicki, fredrik.sandin
        %\}@ltu.se
    }
}

\maketitle

\IEEEpubidadjcol

\begin{abstract}
    \input{Sections/0_abstract.tex}
\end{abstract}


%\IEEEpeerreviewmaketitle

\section{Introduction}
\label{sec:introduction}
\input{Sections/1_introduction.tex}

\section{Materials and Methods}
\label{sec:methods}
\input{Sections/2_methods.tex}

\section{Results and Discussion}
\label{sec:results}
\input{Sections/3_results.tex}

%\section{Discussion}
%\label{sec:discussion}
%\input{Sections/4_discussion.tex}

% Discussion can be sufficient, not necessarily needed:
\section{Conclusion}
\label{sec:conclusion}
\input{Sections/5_conclusion.tex}


%\section*{Acknowledgment}

\bibliographystyle{IEEEtran}
\bibliography{references}

\end{document}

%% file: acronyms.tex
\newacronym{AER}{AER}{Address-Event Representation}
\newacronym{AdEx}{AdEx}{Adaptive Exponential Integrate-and-Fire}
\newacronym{ANN}{ANN}{Artificial Neural Network}
\newacronym{AP}{AP}{Action Potential}

\newacronym{CAM}{CAM}{Content-Addressable Memory}

\newacronym{DPI}{DPI}{Differential Pair Integrator}
\newacronym{DYNAP}{DYNAP}{Dynamic Neuromorphic Asynchronous Processor}

\newacronym{EPSP}{EPSP}{Excitatory Postsynaptic Potential}

\newacronym{FDHM}{FDHM}{Full Duration at Half Minimum}

\newacronym{IPSP}{IPSP}{Inhibitory Postsynaptic Potential}
\newacronym{ISI}{ISI}{Interspike Interval}

\newacronym{PIR}{PIR}{Postinhibitory Rebound}
\newacronym{PSP}{PSP}{Postsynaptic Potential}

\newacronym{SNN}{SNN}{Spiking Neural Network}

%% file: Sections/0_abstract.tex
%
% 1. Area: subject, keywords, relevance
%
Spiking neurons can perform spatiotemporal feature detection by nonlinear synaptic and dendritic integration of presynaptic spike patterns.
%
%
% 2. Problem: Describe the knowledge gap
%
Multicompartment models of nonlinear dendrites and related neuromorphic circuit designs enable faithful imitation of such dynamic integration processes,
but these approaches are also associated with a relatively high computing cost or circuit size.
%
%
% 3. Solution: Describe the solution presented
%
Here, we investigate synaptic integration of spatiotemporal spike patterns with multiple dynamic synapses on point-neurons in the DYNAP-SE neuromorphic processor,
which offers a complementary resource-efficient, albeit less flexible, approach to feature detection.
%
% 4. Methodology: Describe how the problem is solved
%
We investigate how previously proposed excitatory--inhibitory pairs of dynamic synapses can be combined to integrate multiple inputs,
and we generalize that concept to a case in which one inhibitory synapse is combined with multiple excitatory synapses.
%
% 5. Results: Describe explicit/quantitative results
%
We characterize the resulting delayed excitatory postsynaptic potentials (EPSPs) by measuring and analyzing the membrane potentials of the neuromorphic neuronal circuits.
We find that biologically relevant EPSP delays, with variability of order $10$ milliseconds per neuron, can be realized in the proposed manner by selecting different synapse combinations, thanks to device mismatch.
%
% 6. Take Away: Specific - do not speculate/generalize
%
Based on these results, we demonstrate that a single point-neuron with dynamic synapses in the DYNAP-SE can respond selectively to presynaptic spikes with a particular spatiotemporal structure,
which enables, for instance, visual feature tuning of single neurons.

%% file: Sections/1_introduction.tex
% Neuromorphics
Neural circuitry is a main source of inspiration in the development of more efficient and potent computing architectures, such as deep neural networks \cite{LeCun2015}.
The neuron models used in such artificial neural networks are greatly simplified state-based models, which require computationally costly iterations to process the spatiotemporal patterns that characterize most real-world events.
However, the fact that such basic models of neurons are so successfully used in applications motivates further investigations of neuroscientifically inspired computational principles and architectures \cite{Pfeiffer2018deep, Rueckauer2017snn}.

In the quest for more energy- and resource-efficient computing and learning architectures, neuromorphic sensors and processors, which more faithfully reproduce the observed dynamic behavior of neurons, are developed by exploiting the dynamics of conventional microelectronic devices and novel nanomaterials \cite{mead1990neuromorphic,schuman2017survey}.
With such a dynamic computing approach, more resource-efficient signal processing and perception systems can be engineered \cite{indiveri2019space}---one example being event-based vision \cite{gallego2019visionSurvey},
which can benefit applications in real-time interaction systems,
such as robotics and wearable electronics \cite{delbruck2016vision},
for which low power, low latency, and high dynamic range are important properties \cite{liu2019eventDriven}.
%One major problem for such applications is the training of \glspl{SNN}, which, in \cite{samadzadeh2020convolutional}, is addressed with an investigation into spatiotemporal feature extraction with convolutional \glspl{SNN}.

%Biological nervous systems consume less energy than digitally implemented \glspl{ANN}, by orders of magnitude, while also being more flexible and robust \cite{indiveri2019space}.
Dynamic neuromorphic processors have parallel instances of mixed-signal analog/digital circuits, operating in real-time, that emulate the biophysical dynamics of neurons and synapses \cite{Bartolozzi2007synapses,indiveri2011neuromorphic,chicca2014neuromorphic}.
Such processors
%mimic the major information processing aspects of biological circuits, which, from a physical point of view, is
are different in comparison to digital computers, from a physical information-processing point of view.
Consequently, such neuromorphic systems can offer efficiency advantages in the development of computational intelligence inspired by the observed functions of brains, the senses, and neural circuits.

% Neuromorphic Delays
In neuromorphic processing of spatiotemporal patterns, temporal delays are essential computational elements \cite{sheik2013spatio}.
Delays have, for instance, been implemented using dedicated, specifically tuned delay neurons serving as axonal delays in \gls{SNN} architectures \cite{sheik2012emergent, sheik2012exploiting}, as well as using synaptic dynamics \cite{sandin2020synaptic}.
%
% Dendritic Delays
%An analytic method for calculating the time delay and speed of propagation of electrical signals in any passive dendritic tree without the need for numerical simulations, is reported in \cite{agmon1993signal}.
%In \cite{stuart2015dendritic}, the critical role of dendrites and their integration in information processing is underlined.
In biology, the delays of \glspl{EPSP} in dendrites range up to tens of milliseconds \cite{agmon1993signal},
and make out part of the critical role of dendrites in processing of spatiotemporal information in neurons \cite{stuart2015dendritic}.
In neuromorphic systems, dendritic integration has been investigated with nonlinear and multicompartment models---see for example \cite{wang2010multilayer, wang2013active, banerjee2015current, hussain2016morphological, schemmel2017accelerated}.

\begin{figure}[t]
    \centering
    \includegraphics[width=0.7\columnwidth]{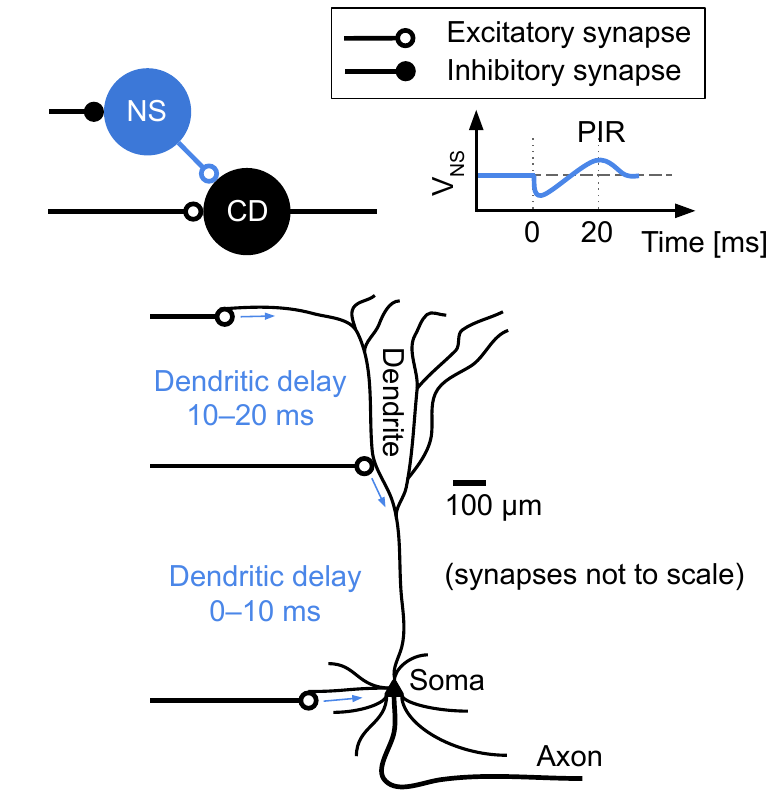}
    \put(-160,170){\textbf{A}}
    \put(-160,110){\textbf{B}}
    \caption
    {
        Examples of feature-selective biological circuits that depend on nonlinear neuronal dynamics.
        \textbf{A:}
        A nonspiking (NS) neuron featuring postinhibitory rebound (PIR) when stimulated by a presynaptic spike-pulse.
        \textbf{B:}
        A pyramidal neuron with millimeter-scale dendrites of varying conductance and capacitance.
    }
    \label{fig:bioInsp}
\end{figure}

% PIR
%The auditory feature detection circuit for sound pattern recognition in the brain of female field crickets described in \cite{schoneich2015auditory} relies on \gls{PIR} in a nonspiking neuron to detect a species-specific sound-pulse interval.
%This circuit has been mimicked in a neuromorphic implementation \cite{arxiv2019disynapticdelays},in which an excitatory--inhibitory pair of synapses was used to imitate the delayed excitation of a coincidence detecting neuron caused by the \gls{PIR} mechanism.
%
Fig.~\ref{fig:bioInsp} illustrates two examples of feature-selective neural circuits based on nonlinear neuronal dynamics.
One such example is illustrated in Fig.~\ref{fig:bioInsp}A,
in which a nonspiking (NS) neuron with one inhibitory synapse is stimulated by a presynaptic spike that leads to a \gls{PIR} of the membrane potential, $V_{NS}$, with maximum after $20$~ms.
The \gls{PIR} generates a delayed \gls{EPSP} in the spiking coincidence-detection (CD) neuron, which implies that the firing probability of the CD neuron depends on the relative timing of presynaptic spikes.
This type of circuit can be observed in the auditory system of crickets \cite{schoneich2015auditory},
and has been mimicked in a neuromorphic implementation \cite{sandin2020synaptic},
in which an excitatory--inhibitory pair of dynamic synapses was used to imitate the delayed excitation of a coincidence detecting neuron caused by the \gls{PIR} mechanism.

A pyramidal neuron with millimeter-scale dendrites of varying width, conductance and capacitance is illustrated in Fig.~\ref{fig:bioInsp}B.
The \glspl{PSP} from excitatory synapses, located (on spines) at different positions along the dendrites, propagate with varying velocity and amplitude depending on the variable properties of the dendrite.
Thus, the propagation of each \gls{PSP} towards the soma is subject to a dendritic delay,
and the relative timing of presynaptic spikes influence their contribution to the eventual firing of the soma, as well as long-term plasticity \cite{stuart2015dendritic}.
Pyramidal neurons are abundant in the neocortex and hippocampus,
and synaptic integration of this type is an essential aspect of neural information processing.
Short-term synaptic plasticity further increases the capacity of synapses and neurons to dynamically integrate temporally encoded information \cite{Buonomano2000decoding,mauk2004temporal}.

%(TODO: what is the knowledge gap? what is your contribution? I am lacking the motivation here.. For example, I think it would be more clear for the reader if we have: problem description, what has been done so far/relative work, what is the gap, what part of the gap we are doing in this paper. )

%emulate different spike types observed in cortical pyramidal neurons: NMDA plateau potentials, calcium and sodium spikes.
% plasticity mechanisms can modify not only the synaptic weights, but also the dendritic synaptic composition

% background, implementing axonal delays efficientyly in neuromorphic hardware
% https://www.ncbi.nlm.nih.gov/pmc/articles/PMC3272652/
% https://ieeexplore.ieee.org/document/6252636

% background, synaptic/dendritic integration
% Sprustons papers, general
% Shih-Chiis papers on multicompartment hardware, https://doi.org/10.1162/neco.2010.06-09-1030
% More recent papers by people with background at INI, Mattias to insert references
% Multi-compartment extension to the BrainScaleS neuron circuits, https://arxiv.org/pdf/1703.07286.pdf

% \cite{
% A,B % INI original contributions
% C, % INI state of art etc
% }

Neuromorphic multicompartment models enable increasingly faithful and flexible implementations of dendritic integration and plasticity \cite{wang2010multilayer,wang2013active,hussain2016morphological, schemmel2017accelerated}.
However, such neuromorphic circuits are also complex, and require larger neuromorphic circuit designs and more power than dynamic point-neuron implementations,
which can matter in resource-constrained applications with high-dimensional inputs, such as battery-powered machine vision systems.
The results in \cite{sandin2020synaptic}, in which excitatory--inhibitory pairs of dynamic synapses on point neurons are used to generate a delayed \gls{EPSP}, suggest that multiple dynamic synapses of that type can potentially be used to integrate spatiotemporal spike patterns within single point-neurons in a dynamic neuromorphic processor.
To what extent can patterns with different temporal extensions and spatial dimensions be detected that way?

Here, we investigate synaptic integration of spatiotemporal spike patterns with multiple dynamic synapses \cite{Bartolozzi2007synapses} on point neurons in the DYNAP-SE neuromorphic processor \cite{moradi2018dynaps}.
The DYNAP-SE is a mixed-signal processor for low-power, real-time emulation of \glspl{SNN},
providing a platform for spike-based neural processing with colocalized memory and computation \cite{indiveri2015pieee}.
We characterize the resulting delayed \glspl{EPSP} by measuring and analyzing the membrane potentials of the neuromorphic neuron circuits,
and we find that biologically relevant \gls{EPSP} delays with variability of order $10$ milliseconds per neuron can be realized.
Albeit less flexible and general than a multicompartment implementation,
our presented work offers a complementary resource-efficient approach to integration and detection of spatiotemporal features.

The contribution of this work is twofold:
(i) we use dynamic synapses in the DYNAP-SE neuromorphic processor integrating multiple delayed \glspl{EPSP} as a simple model of dendritic integration \cite{wang2013active, schemmel2017accelerated};
(ii) we model, in effect, axonal, as well as dendritic and synaptic, temporal delays---instead of only axonal ones \cite{sheik2012emergent, sheik2012exploiting}---and we thereby perform synaptic integration of spatiotemporal information using point neurons in a mixed-signal neuromorphic processor, which is subject to device-mismatch related challenges.

%To what extent can dendritic/synaptic integration be implemented in a dynamic neuromorphic processor using synaptic dynamics subject to device mismatch, without explicit multicompartment circuits?

%Related Work
%
%Sheik et al. generated a range of temporal delays, exploiting device mismatch---STDP was used on the Exc connections from the delay neurons \cite{sheik2012emergent, sheik2012exploiting, coath2014robust}.
%
%Wang et al. used \textit{delay adaptation}---inspired by STDP---to learn delays for a polychronous network \cite{wang2013fpga, wang2014mixed}.
%
%In \cite{wang2010multilayer}, a real-time analog VLSI dendritic compartment model is used to investigate the neuronal processing of temporal input patterns on the dendrites.

% pyramidal neurons are the most abundant cells of the neocortex
% synaptic integration in dendrites 
% fundamental for cortical tasks such as feature selection

% Figure 1
% https://docs.google.com/drawings/d/1g2t-b-sdeXFw7ouSDJPsLo6RvGrntXAw3VUqtDEFYRY/edit?usp=sharing

%% file: Sections/2_methods.tex
%\cite{Nirenberg7348}

% Forward search on Nirenberg7348 gives eg
% https://www.nature.com/articles/nn.2842
% https://www.nature.com/articles/nrn2578
% https://www.nature.com/articles/nn.3330
% https://www.jneurosci.org/content/30/13/4815.short
% https://journals.plos.org/ploscompbiol/article?id=10.1371/journal.pcbi.1000205
% https://journals.plos.org/ploscompbiol/article?id=10.1371/journal.pcbi.1004083
% https://www.jneurosci.org/content/25/21/5195.short
%       PDF, see Eq 5: https://www.jneurosci.org/content/jneuro/25/21/5195.full.pdf
% Novel concepts: https://link.springer.com/content/pdf/10.1007%2F978-3-642-53734-9.pdf

% Experimental Setup
The experimental setup used in this work consisted of a DYNAP-SE unit---a \gls{DYNAP} \cite{moradi2018dynaps} from SynSense---connected to a PC via a USB interface.
The DYNAP-SE was controlled from the PC using the cAER event-based processing framework for neuromorphic devices.
Since the DYNAP-SE emulates neuronal and synaptic dynamics in real-time---using analog circuitry---we configured it in a hardware-in-the-loop setup,
in which a PC receives digital spike-event data and analog neuron monitoring,
while iteratively reconfiguring the DYNAP-SE in order to carry out the measurement series described in the following.
All stimuli were synthetically generated using the built-in FPGA spike-generator in the DYNAP-SE,
which generates spike-events according to assigned \glspl{ISI} and virtual source-neuron addresses.
The 8-bit USB oscilloscope SmartScope from LabNation was used for measurements of analog neuronal membrane potentials in the DYNAP-SE.

\subsection{The DYNAP-SE Neuromorphic Processor}
\label{sec:dynap-se}

\begin{figure}[t]
    \centering
    \includegraphics[width=\columnwidth]{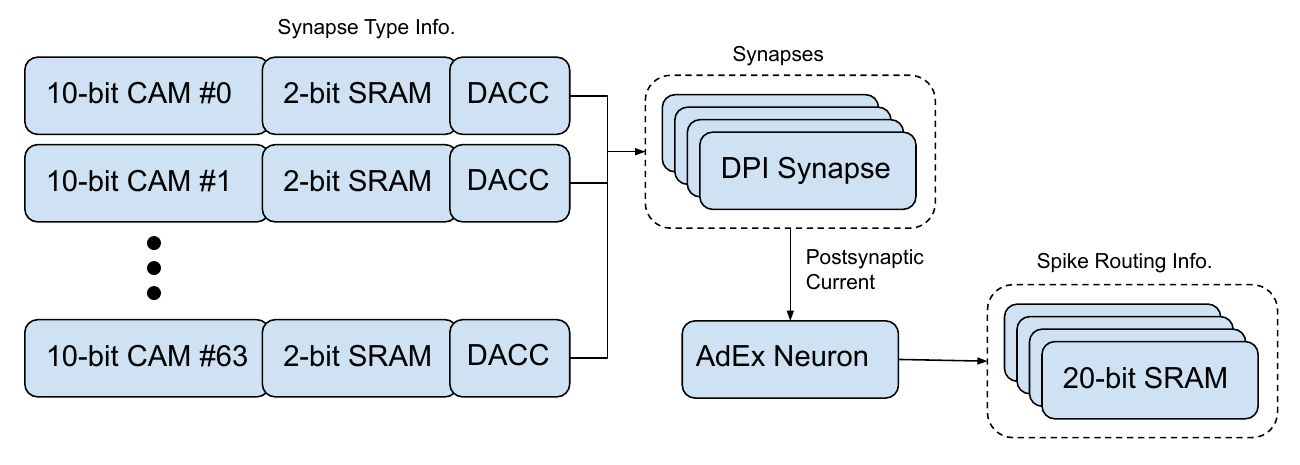}
    %\put(-100,100){\textbf{A}}
    \caption
    {
        Simplified block diagram of one of the 256 %computational 
        mixed-signal analog/digital
        neuronal nodes in each core of the DYNAP-SE.
        Each node contains 64 mixed-memory words, each with a 10-bit CAM cell and a 2-bit SRAM cell,
        four synaptic DPI circuits,
        and one AdEx neuron circuit.
        The digital-to-analog signal-converting circuitry for each mixed-memory word is here simplified with a block labeled Digital-to-Analog Current Converter (DACC).
        Also, the four 20-bit SRAM cells holding fan-out spike routing information are displayed.
    }
    \label{fig:dynapNode}
\end{figure}

% DYNAP-SE Overview
The DYNAP-SE is a reconfigurable, general-purpose, mixed-signal \gls{SNN} processor,
which uses low-power, inhomogeneous, sub-threshold, analog circuits to emulate the biophysics of neurons and synapses in real-time.
One DYNAP-SE unit comprises four four-core chips---each core having 256 \gls{AdEx} neuron-circuits.
Each neuron has a \gls{CAM} block containing 64 addresses, see Fig.~\ref{fig:dynapNode},
which represent connections to presynaptic neurons.
Four different synapse types are available for each connection:
fast and slow excitatory, and subtractive and shunting inhibitory, respectively.
The dynamic behaviors of the neuronal and synaptic circuits in the DYNAP-SE are governed by analog circuit parameters,
which are set by programmable on-chip bias-generators providing 25 bias parameters independently for each core.
Information about spike-events is transmitted between the neurons of the DYNAP-SE using the \gls{AER} communication protocol.

\subsubsection{Spiking Neuron Model}

% AdEx Neuron Model
The \gls{AdEx} spiking neuron model \cite{brette2005adaptive} describes the neuronal membrane potential, $V$, and an adaptation variable, $w$, with two coupled nonlinear differential equations:
\begin{subequations}
    \begin{equation}
	    C\frac{dV}{dt}=-g_{L}(V-E_{L})+g_{L}\Delta_{T}e^{\left( V-V_{T}\right)/\Delta_{T}}-w+I,
		\label{eq:AdEx_memPot}
	\end{equation}
	\begin{equation}
		\tau_{w}\frac{dw}{dt}=a\left(V-E_{L}\right)-w,
		\label{eq:AdEx_adapt}
	\end{equation}
	\label{eq:AdEx}
\end{subequations}
in which
$C$ is the membrane capacitance,
$g_L$ the leak conductance,
$E_L$ the leak reversal potential,
$V_T$ the spike threshold,
$\Delta_T$ the slope factor,
$I$ the postsynaptic input current,
$\tau_w$ the adaptation time constant, and
$a$ the subthreshold adaptation.
For $V > V_T$, the membrane potential increases rapidly, due to the nonlinear exponential term,
leading to a rapid depolarization and spike generation,
at time of which, $t = t_{spike}$, the membrane potential and the adaptation variable are both, respectively, updated according to
\begin{subequations}
	\begin{equation}
		V \rightarrow V_{r},
		\label{eq:AdEx_voltRes}	
	\end{equation}
	\begin{equation}
		w \rightarrow w+b,
		\label{eq:spikeTrig_Adapt}	
	\end{equation}
	\label{eq:AdEx_spike}
\end{subequations}
where $V_r$ is the neuronal reset potential and $b$ is the spike-triggered adaptation.

\subsubsection{Dynamic Synapse Model}
% Synapse Model
The synapses of the DYNAP-SE are implemented with subthreshold \gls{DPI} log-domain filters,
which are proposed in \cite{Bartolozzi2007synapses} and further described in \cite{chicca2014neuromorphic}.
The following first-order linear differential equation approximates the response of a \gls{DPI} to an input current $I_{in}$:
\begin{equation}
	\tau\frac{d}{dt} I_{out} + I_{out} = \frac{I_{th}}{I_{\tau}}I_{in},
    \label{eq:DPI}
\end{equation}
where 
$I_{out}$ is the postsynaptic output current,
$\tau$ and % the time constant,
$I_{\tau}$ are time-constant parameters, and
$I_{th}$ is an additional control parameter that can be used to change the gain of the filter.
This approximation is valid for $I_{in} \gg I_{\tau}$ and $I_{out} \gg I_{I_{th}}$.

\subsection{Disynaptic Delays}
\label{sec:synDelays}

% Disynaptic Delays  \todo{ all 4 different synapses types or some of them?} [x]
We used excitatory--inhibitory pairs of dynamic synapses in the DYNAP-SE to implement temporally delayed interneuronal connections in the DYNAP-SE---in the manner that is described in detail in \cite{sandin2020synaptic}.
More specifically, one excitatory--inhibitory synapse pair, connected to the same input-neuron, constitutes one delay element and---in a manner resembling \gls{PIR}---generates a delayed excitation in the postsynaptic neuron upon stimulation.
For the inhibition, a synapse of the subtractive type was used,
which allows the combination of excitation and inhibition by summation of the postsynaptic currents.
A synapse of the slow type was used for the excitation, which operates with on a relatively long time-scale---leaving the fast type available for use for direct stimulation of the neuron, in potential future cases.
The excitation delay was realized by giving the excitatory synapse a longer time-constant than that of the inhibitory one,
so that, following the decay of the inhibition---which was set to a time-constant matching the desired temporal delay---the \gls{EPSP} still contributes to raise the neuronal membrane potential and, thereby, generates the delayed excitation.
The bias-parameter values used for this configuration of the DYNAP-SE are provided in Table~\ref{tab:biasVals}.
\begin{table}[t]
    \caption
    {
        Bias parameter values used to implement disynaptic delay elements in the DYNAP-SE.
    }
    \centering
    \begin{tabular}{l l c c l}
        \toprule
        \textbf{Parameter type} & \textbf{Parameter name} & \textbf{Coarse} & \textbf{Fine} & \textbf{Current} \\
        & & \textbf{value} & \textbf{value} & \textbf{level} \\
        \midrule
        Neuronal
        & \texttt{IF\_AHTAU\_N} 			& 	7   & 	35  &   L   \\
        & \texttt{IF\_AHTHR\_N} 			& 	7 	& 	1   &   H   \\
        & \texttt{IF\_AHW\_P} 				&	7 	& 	1 	&   H   \\
        & \texttt{IF\_BUF\_P}				& 	3	& 	80	&   H   \\
        & \texttt{IF\_CASC\_N}				& 	7	& 	1	&   H   \\
        & \texttt{IF\_DC\_P}				&	1	&	30	&   H   \\
        & \texttt{IF\_NMDA\_N}				& 	1	& 	213	&   H   \\
        & \texttt{IF\_RFR\_N}				&	4	& 	40	&   H   \\
        & \texttt{IF\_TAU1\_N}				& 	5	& 	39	&   L   \\
        & \texttt{IF\_TAU2\_N}				& 	0	&	15	&   H   \\
        & \texttt{IF\_THR\_N}				&	6	& 	135	&   H   \\
        \midrule
        Synaptic
        & \texttt{NPDPIE\_TAU\_S\_P}		& 	5	& 	70	&   H   \\
        & \texttt{NPDPIE\_THR\_S\_P}		& 	0 	&	210	&   H   \\
        & \texttt{NPDPII\_TAU\_F\_P}		& 	5	& 	100	&   H   \\
        & \texttt{NPDPII\_THR\_F\_P}		& 	3	& 	60	&   H   \\
        & \texttt{PS\_WEIGHT\_EXC\_S\_N}	& 	0	& 	140	&   H   \\
        & \texttt{PS\_WEIGHT\_INH\_F\_N}	& 	0	& 	150 &   H   \\
        & \texttt{PULSE\_PWLK\_P}			&	5	& 	40	&   H   \\
        & \texttt{R2R\_P}					& 	4	& 	85	&   H   \\
        \bottomrule
    \end{tabular}
    \label{tab:biasVals}
\end{table}

\begin{figure}[t]
    \centering
    \includegraphics[width=0.5\columnwidth]{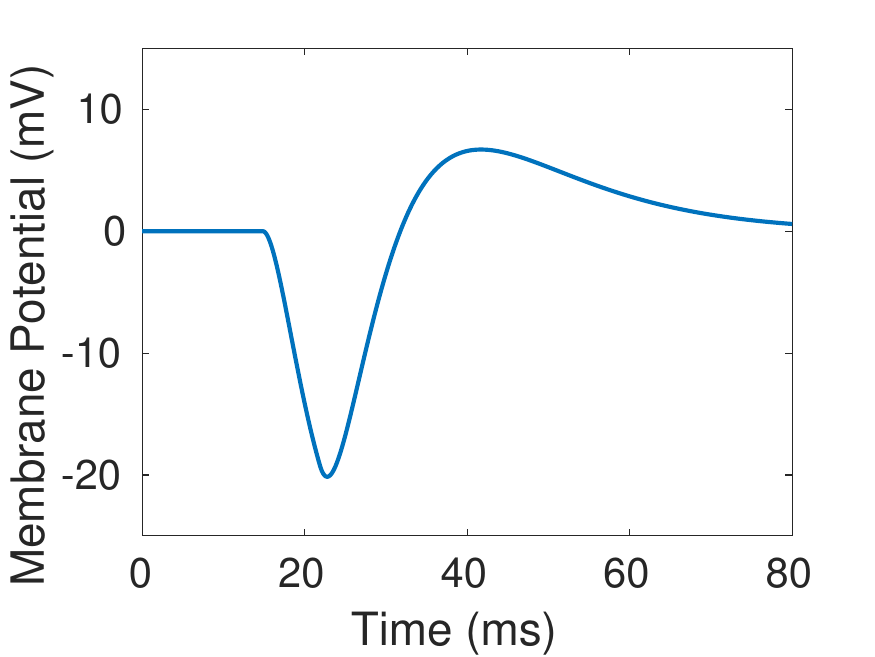}
    % \put(-100,100){\textbf{A}}
    \caption
    {
        Simulation of the disynaptic delay-element model \cite{sandin2020synaptic}.
        The figure shows the postsynaptic neuronal membrane potential following presynaptic single-spike stimulation of one delay element.
    }
    \label{fig:delaySim}
\end{figure}

% Simulation of Delay Elements
The disynaptic delay elements can be simulated using Eq.~(\ref{eq:DPI}),
and the postsynaptic neuronal membrane potential using Eq.~(\ref{eq:AdEx}).
Fig.~\ref{fig:delaySim} shows the result of such a numerical simulation for a single-spike input.
Since the simulated neuron is in the subthreshold regime, where $V < V_{T}$,
Eq.~(\ref{eq:AdEx}) was simplified by setting the exponential term to zero, and by omitting the adaptation variable.
The neuronal and synaptic parameters used in the simulation were selected for the neuronal membrane potential to be comparable to those measured in the DYNAP-SE,
and should, therefore, not be directly compared with potentials and threshold values.

% Distribution of Delays
Due to the device mismatch inherent to the analog neuronal and synaptic circuits of the DYNAP-SE,
any set of bias-parameter values generates a distribution of the corresponding neuronal and synaptic dynamic behaviors in the core being configured.
Thus, implementation of the disynaptic delays as described above---by configuring the bias parameters of one core of the DYNAP-SE accordingly---should generate a distribution of delays in the population of neurons.
Furthermore, even though all 64 \glspl{CAM} of one DYNAP-SE neuron technically use the same four synaptic circuits---for the four different synapse types, respectively---there is digital-to-analog current-converting circuitry between the \glspl{CAM} and the synaptic circuits,
which constitutes a further source of inhomogeneity.
Thus, different disynaptic delays implemented on the same neuron, but using different \glspl{CAM}, are expected to exhibit some degree of variation in behavior,
why a distribution of temporal delays can be expected also in one and the same neuron.

\subsection{Feature Detection Architectures}
\label{sec:featDet_archs}

% Feature-Detection Intro.
Given the expected temporal variation in disynaptic delay elements implemented using different \glspl{CAM} on the same single neuron,
input-spikes arriving to such a neuron---via different delay elements---should generate \glspl{EPSP} with coincident maxima,
if the differences in presynaptic spike times compensate for the differences in the synaptic delay times.
Thus, input patterns with spike-time intervals that match the delay-time differences should generate maximal excitation of the neuron,
why a single neuron should be able to respond selectively---with increased intensity---to such spatiotemporal input patterns.
To investigate this, we performed two different experiments,
in which single neurons were set up to receive spatiotemporal input spike-patterns consisting of temporally separated single spikes received through different input channels.
In both of the experiments---described in the following---an off-line Hebbian-like learning rule was used to select the synapses of the neurons, for them to respond selectively to different \glspl{ISI} in the input spike-patterns.
More specifically, we investigated whether the single-neuron systems could respond with increased intensity to some limited range of pattern \glspl{ISI} in the millisecond-range,
and, thereby, discriminate against both longer and shorter intervals.

\subsubsection{Pair-Selective Circuit}

We configured a single neuron with two inputs via two different excitatory--inhibitory disynaptic delay elements configured as described in Section~\ref{sec:synDelays}.
%The single neuron was set to receive two spike-inputs of two different excitatory synapses.
The input pattern consisted of a pair of spikes separated with an \gls{ISI}---one spike to each delay element (see Section~\ref{sec:pairDet_results}).
The delay-element synapses were selected for the neuron to respond selectively to intermediately long intervals but not to shorter or longer intervals.

\subsubsection{Triplet-Selective Circuit}

% Triplet-Detection
To investigate the generalizability of our use of synaptic dynamics for single-neuron spatiotemporal pattern recognition,
we set up a single neuron to receive single-spike inputs on three different excitatory synapses and one inhibitory synapse.
In this experiment, the stimulation pattern consisted of one single spike to each of the excitatory synapses, each spike temporally separated from the previous one with the same \gls{ISI}---such that the first and the third spike were separated with twice the \gls{ISI}---as well as one spike to the inhibitory synapse, simultaneous with the first excitatory spike (see Section~\ref{sec:tripDet_results}).
The same bias parameter values as in the pair-detection experiments were used,
except for a lowering of the excitatory synaptic weight---to compensate for the higher number of excitatory synapses and lower number of inhibitory ones.
Synapses were selected for the neuron to respond with increased intensity to a range of intermediately long \glspl{ISI}, as compared to shorter and longer intervals.

\begin{figure}[t]
    \centering
    \begin{subfigure}[b]{0.85\columnwidth}
        \includegraphics[width=\textwidth]{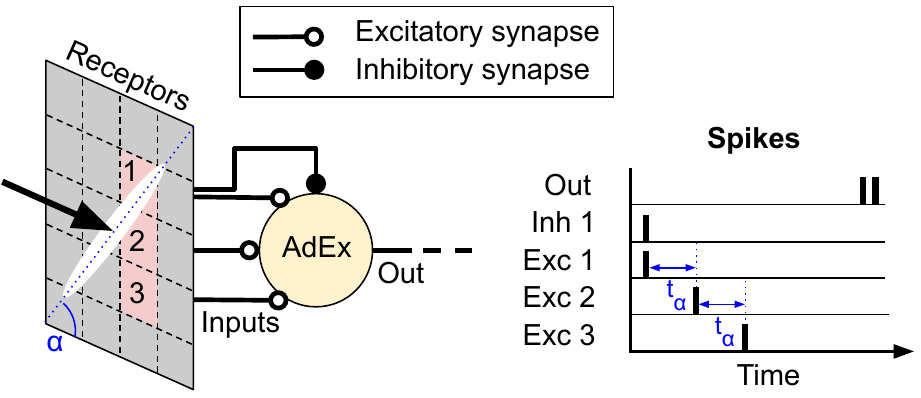}
    \end{subfigure}
    \put(-180,85){\textbf{A}}
    \put(-20,65){\textbf{B}}
    \\
    \begin{subfigure}[b]{0.49\columnwidth}
        \includegraphics[width=\textwidth]{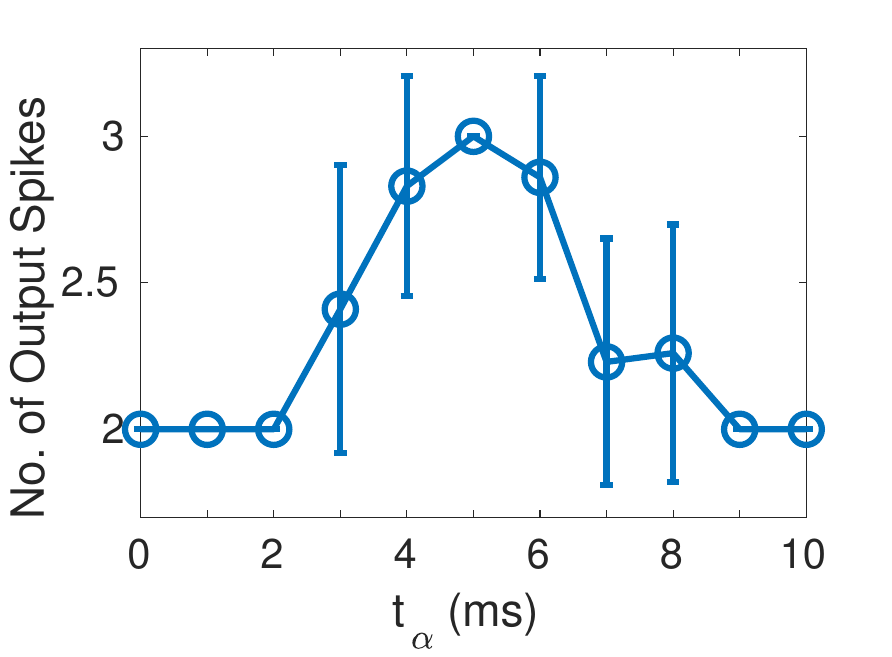}
    \end{subfigure}
    \begin{subfigure}[b]{0.49\columnwidth}
        \includegraphics[width=\textwidth]{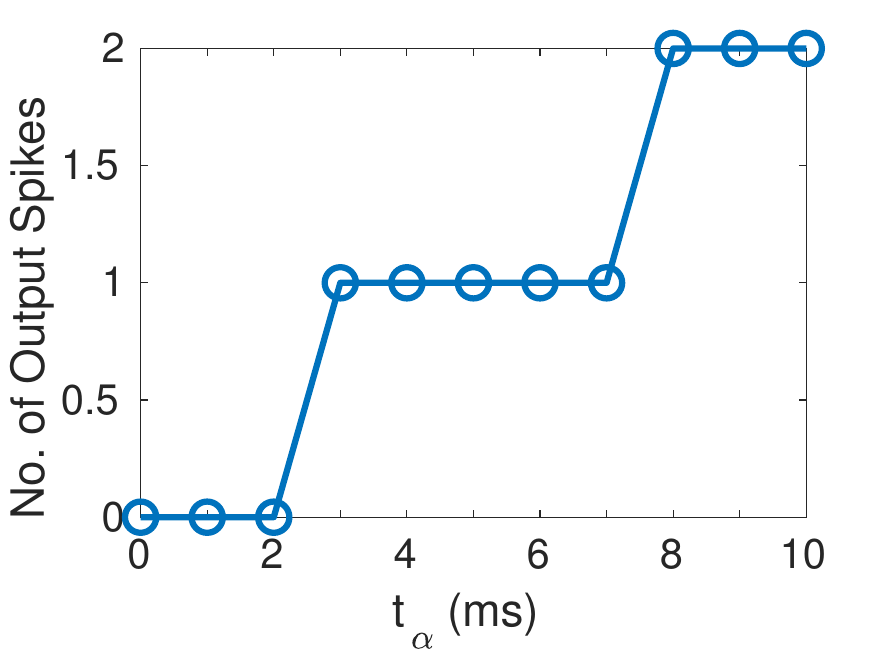}
    \end{subfigure}
    \put(-220,70){\textbf{C}}
    \put(-95,70){\textbf{D}}
    \caption
    {
        Selective response of a single hardware neuron with dynamic synapses to different visual stimuli.
        % input ISIs in spatiotemporal spike-triplet feature.
        \textbf{A:}
        One hardware AdEx neuron receives inputs from three simulated visual receptors (red squares) that output spikes asynchronously when detecting a contrast change.
        %A bright stimulus (white ellipsis) with angle $\alpha$ moves across the receptor array, causing the three receptors to fire a sequence of spikes.
        %Three excitatory synapses and one inhibitory synapse are configured to make the neuron fire selectively to stimuli with a preferred orientation and speed.
        \textbf{B:}
        Spikes resulting from the presentation of one visual stimulus.
        The three excitatory synapses receive, respectively, one presynaptic spike from each receptor,
        with time difference $t_\alpha$ that depends on the orientation and speed of the stimulus.
        The inhibitory synapse receives a presynaptic spike from one of the three receptors (inhibitory interneuron not required in DYNAP processors).
        %Depending on the timing between spikes, $t_\alpha$, this neuron fires 2--3 output spikes for each presented stimulus.
        \textbf{C:}
        %Average number of output spikes per stimulus (N = 100) versus the time between spikes, $t_\alpha$.
        Selective response (N = 100) of a single DYNAP-SE neuron to the spatiotemporal spike-triplet stimulus illustrated in panels A and B, for different values of the time-interval $t_\alpha$.
        Error bars denote $\pm$1 standard deviation.
        \textbf{D:}
        Response (N = 100) to spike-triplet vs $t_{\alpha}$ for a different neuron and selection of synapses---the standard deviation is zero for all data-points in this case.
        Due to device mismatch, the feature tuning curves are neuron- and synapse-specific.
    }
    \label{fig:tripletDet}
\end{figure}

% Triplet Application
The stimulation pattern used in this experiment can be likened to the response of three spatially distributed contrast-detecting visual receptor neurons to a bright line moving across the visual field of the receptor array, causing each receptor to fire asynchronously; this concept is illustrated in Fig.~\ref{fig:tripletDet}.
% Figure moved here to maintain the correct ordering of figure references. It also includes a pedagogic illustration of the experiments carried out, which may help the reader to understand the more abstract subsequent plots
%https://docs.google.com/drawings/d/1Rl6CxJVlOe8XXMSeS1gMeEz8katyjPwqVYd898jdcP8/edit?usp=sharing
%
This setup is aligned with the fact that biological vision is highly sensitive to contrast changes rather than to the overall illumination,
and that a neuromorphic vision system such as that in \cite{Brandli2014dvs} would generate an output of this type.
As a historical note, in 1981, Hubel and Wiesel \cite{hubel1959receptive} got the Nobel Prize in Psychology for their discoveries concerning the visual system.
In their experiment, they used the projection of a single line in different orientations as stimulus,
while they were recording the activity of a single neuron in the cats brain.
They discovered that the specific neuron was highly activated when then line had a vertical orientation.

In the example described above, both the angular orientation and the velocity of the stimulus would influence the \gls{ISI} separating the asynchronous responses of the receptor cells.
%Furthermore, it is the receptor neuron that is expected to fire first in the case of feature detection that has an inhibitory connection---in addition to its excitatory one---to the feature detecting neuron.
Furthermore, the projection to the inhibitory synapse, as well as the specific \gls{EPSP} delays of the excitatory synapses, determines the feature tuning of the neuron.
This solution is possible because, in the DYNAP architecture---as opposed to in biology---inhibitory interneurons are not required.

%% file: Sections/3_results.tex
\subsection{Delay Characteristics}

\begin{figure}[t]
    \centering
    \begin{subfigure}[b]{0.49\columnwidth}
        %\caption{}
        \includegraphics[width=\textwidth]{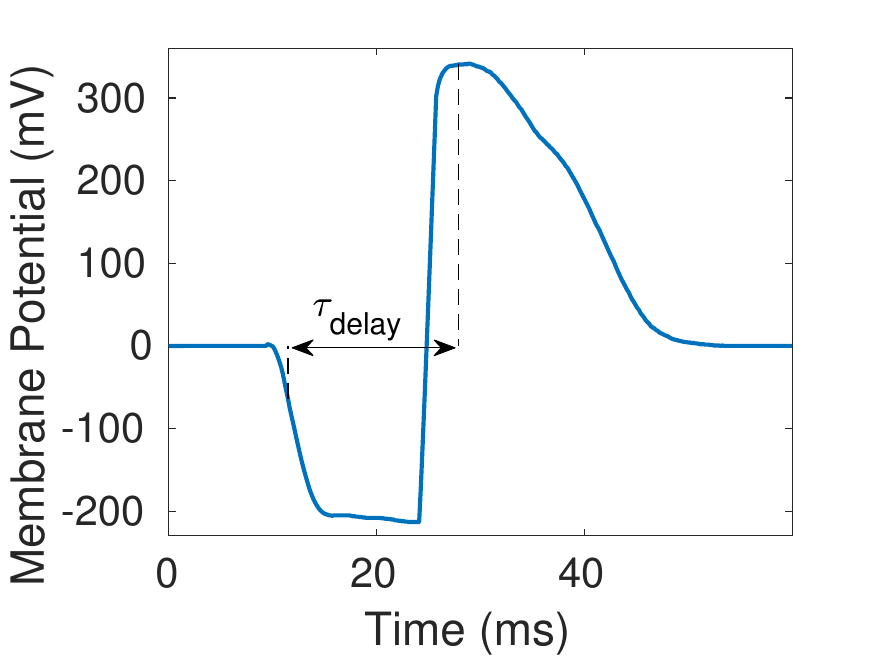}
    \end{subfigure}
    \put(-95,90){\textbf{A}}
    \\
    \begin{subfigure}[b]{0.49\columnwidth}
        %\caption{}
        \includegraphics[width=\textwidth]{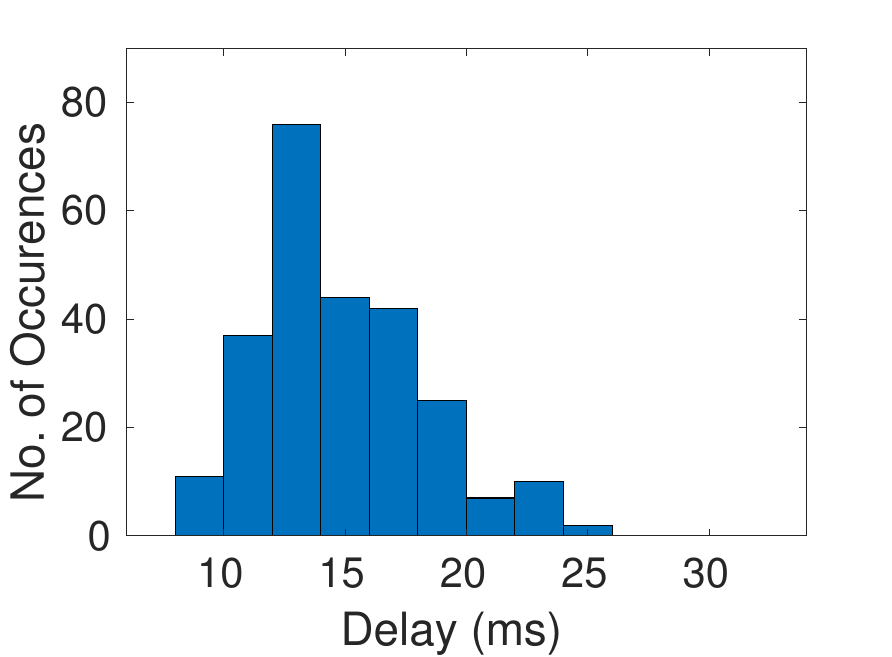}
    \end{subfigure}
    \put(-95,90){\textbf{B}}
    \begin{subfigure}[b]{0.49\columnwidth}
        %\caption{}
        \includegraphics[width=\textwidth]{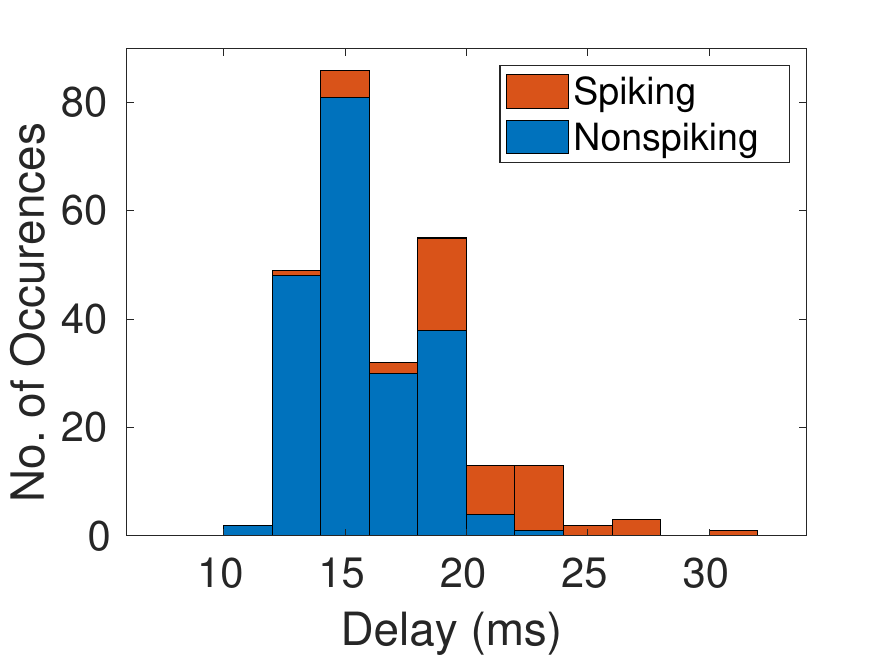}
    \end{subfigure}
    \put(-95,90){\textbf{C}}
    \caption
    {
        Characteristics of disynaptic delay elements implemented in the DYNAP-SE neuromorphic processor, for single-spike input.
        \textbf{A:}
        Subthreshold (nonspiking) neuronal membrane potential for single-spike stimulation via one disynaptic delay element.
        \textbf{B:}
        Distribution (N = 256) of temporal delays when implemented on different neurons with shared global biases.
        \textbf{C:}
        Distribution (N = 256) of temporal delays when implemented on the same neuron, but using different synapse-CAM combinations.
        The two colors represent CAM configurations for which the neuron was spiking or nonspiking, respectively.
    }
    \label{fig:delayHist}
\end{figure}

% Population Delay-Distribution
We implemented the disynaptic delays, as described in Section~\ref{sec:synDelays}, in parallel, on all neurons in one core of the DYNAP-SE---one delay element on each neuron (N=256).
For the purpose of characterization, we defined the duration of the delay as spanning from the onset of the inhibition to the maximum postinhibitory value of the membrane potential---making the definition practical also for neurons that generate a spike as a consequence of their delayed excitation.
The onset of the inhibition was defined at half minimum of the membrane potential, in line with the definition of \gls{FDHM}, given the lack of exact spike-time data in the analog measurements.
Fig.~\ref{fig:delayHist}A shows the membrane potential, following a single-spike input stimulus, of a neuron from the configured core, alongside an illustration of the temporal delay.
While this neuron exhibits typical behavior in the nonspiking case,
spike-firing was triggered by the delayed excitations in roughly half of the neurons in the population.
The resulting distribution of temporal delays is presented in Fig.~\ref{fig:delayHist}B,
according to which, for example, almost 80 out of the 256 neurons display an \gls{EPSP} that is delayed by about 15~ms.

% Single-Neuron Delay-Distribution
Furthermore, we characterized the distribution of temporal delays that are generated in a single neuron when varying the \glspl{CAM} used for the two synapses that constitute one delay element.
We did this by configuring the disynaptic delay in 256 different instances,
using unique pairs of \glspl{CAM} each time.
The resulting delay distribution is presented in Fig.~\ref{fig:delayHist}C.
The bimodal shape of this histogram appears because some of the longer delays correspond to \gls{CAM} combinations where the neurons spikes, while others do not.
When the neuron spikes, the duration of the spike-firing process adds to the delay, according to the delay definition used here.
The peak at shorter delay corresponds to nonspiking instances,
and the second peak is formed where the largest number of spiking delays are found.

\subsection{Spike-Pair Selectivity}
\label{sec:pairDet_results}

\begin{figure}[t]
    \centering
    \begin{subfigure}[b]{0.49\columnwidth}
        \includegraphics[width=\textwidth]{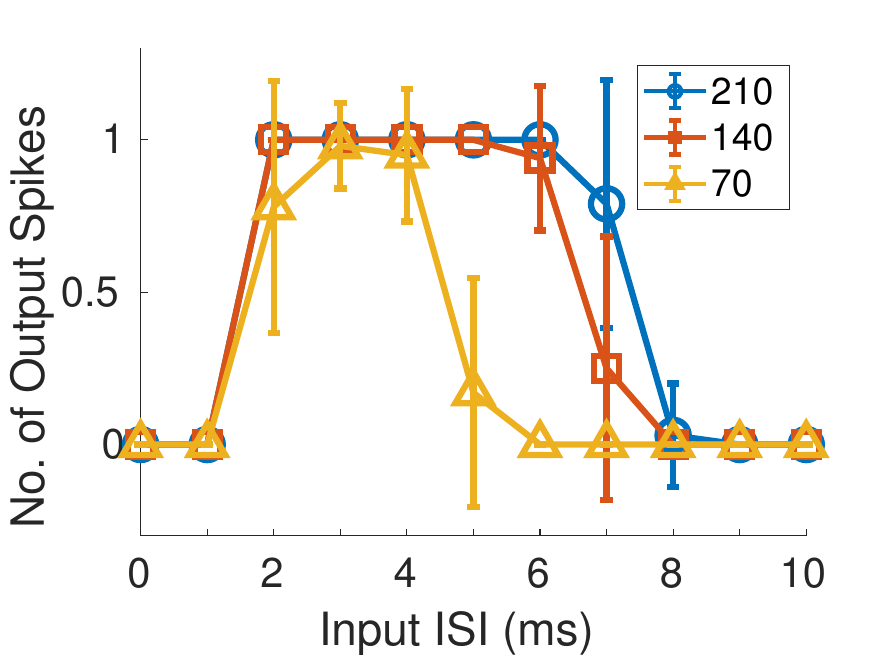}
    \end{subfigure}
    \put(-100,95){\textbf{A}}
    \begin{subfigure}[b]{0.49\columnwidth}
        \includegraphics[width=\textwidth]{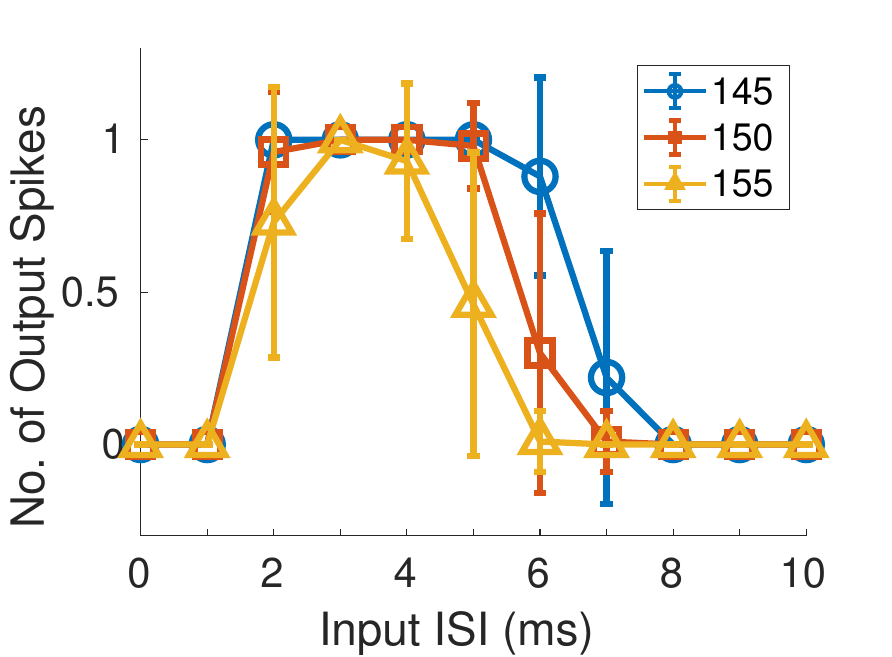}
    \end{subfigure}
    \put(-100,95){\textbf{B}}
    \caption
    {
        Selective responses (N = 100) of a single neuron to different input ISIs in spatiotemporal spike-pair feature for different excitatory and inhibitory synaptic weights, respectively.
        The legends denote the fine integer value of the bias parameter corresponding to the varied weight.
        Error bars denote $\pm$1 standard deviation.
        \textbf{A:}
        Varying excitatory synaptic weight.
        \textbf{B:}
        Varying inhibitory synaptic weight.
    }
    \label{fig:pairDet}
\end{figure}

% Pair-Detection Results
We stimulated the spike-pair sensitive neuron described in Section~\ref{sec:featDet_archs} with \glspl{ISI} ranging from 0 to 10~ms, with increments of 1~ms.
Stimulation with each \gls{ISI} was repeated 100 times, in order to extract the mean number of spikes generated in the receiving neuron.
This investigation was repeated with variations to both the excitatory and the inhibitory synaptic weights of the delay elements,
through which the neuron received the stimuli.
The results, presented in Fig.~\ref{fig:pairDet}, show that the neuron responded selectively to the different \glspl{ISI},
and how this selectivity varies for different choices of the synaptic weights.
Summing the data-points for each of these tuning curves, respectively, generates the following relative areas under the curves:
5.87,
5.19,
and 2.88
for Fig.~\ref{fig:pairDet}A,
and
5.10,
4.25,
and 3.13,
for Fig.~\ref{fig:pairDet}B.

\subsection{Spike-Triplet Selectivity}
\label{sec:tripDet_results}

% Triplet-Selectivity
In the investigation of triplet-interval sensitivity, as in the pair-selection experiment, we stimulated the neuron with \glspl{ISI} of 0 to 10~ms, with increments of 1~ms---meaning that, for the largest \gls{ISI}, the first spike and the third were, on different input synapses, separated with 20~ms.
The results, presented in Fig.~\ref{fig:tripletDet}, illustrates that the neuron---having a baseline response of two spikes per input---does indeed respond with increased average activity for input \glspl{ISI} ranging between 3 and 8~ms.
The response peaks at three spikes per stimulus for the 5-ms \gls{ISI}.
This selective response disappeared when we permuted the order of the excitatory synapses---as is expected, since maximum excitation is obtained when the delayed \glspl{EPSP} are matched by the timings of the presynaptic spikes.

% Discussion: Triplet Variation
One slightly more complex variation to the spike-triplet feature could consist of three spikes separated by two different \glspl{ISI} instead of the uniform interval $t_{\alpha}$.
This would represent stimulation by a nonlinear visual edge or, alternatively, by the same linear edge as in Fig.~\ref{fig:tripletDet}A,
except now presented to the receptors with a nonuniform speed.
Such a feature could be detected using the same architecture as described in this work---simply requiring a different choice of synapse-\glspl{CAM},
for the \glspl{EPSP} to add up maximally.

% Discussion: Increased Complexity
For learning and detection of more complex spatiotemporal features and patterns, more than one neuron is, of course, necessary---see for instance \cite{Pfeiffer2018deep, samadzadeh2020convolutional} for further reading.
One immediate step from the work presented in this paper could be to combine the tuning curves of two or more of the proposed single-neurons systems,
in order to---where these tuning curves correlate most strongly---create a narrower tuning curve in a subsequent neuron.
Activation of this neuron would, then, depend on the feature sensitivities of the preceding neurons,
and the relative timing with which these are activated.

% E.g., with different ISIs - t_α1 and t_α2 – 1) would require a different combination of synapses, 2) or a population of neurons if the non-linear features are more complex. The finite width of the tuning curve is a benefit in that case. 

%% file: Sections/5_conclusion.tex
% Conclusion
In this paper, we propose a resource-efficient approach to spatiotemporal pattern recognition using dynamic synapses and point-neurons in the DYNAP-SE neuromorphic processor to, in effect, model axonal delays and some aspects of dendritic integration.
We use this approach to integrate multiple inputs by using excitatory--inhibitory disynaptic delay elements \cite{sandin2020synaptic}.
Furthermore, we generalize this concept by combining one inhibitory synapse with multiple excitatory synapses.
We conclude that biologically relevant \gls{EPSP} delays with a variability in the order of 10~ms per neuron can be realized due to device mismatch in the analog electronic neuromorphic circuits.
Based on these findings, we demonstrate that a single point-neuron with dynamic synapses in the DYNAP-SE can respond selectively to presynaptic spikes with a particular spatiotemporal structure,
which enables feature detection with single neurons.
We note that the temporal feature tuning of the neuromorphic neurons, as illustrated in Fig.~\ref{fig:tripletDet}C, is comparable to the width of temporal feature detection neurons in biology, see for example Fig.~3B in \cite{schoneich2015auditory}.
Further work is required to investigate how \glspl{SNN} with feature detectors of this type could be configured and trained in a systematic manner given a particular task,
in order to make efficient use of the dynamic synapses.
Further work is also required to investigate under what conditions a simple and relatively resource-efficient feature detector of this type is favored over a more generic multicompartment model of nonlinear dendrites.